\documentclass[12pt]{article}
\usepackage{graphicx,color}
\usepackage{natbib}
\usepackage{amsfonts}
\usepackage{mathtools}
\usepackage{amsthm,amsmath,natbib,algorithm,algpseudocode,pifont,lscape,varwidth}
\usepackage[mathscr]{euscript}
\usepackage{subcaption,enumitem}
\usepackage{bm}
\usepackage{authblk} 
\usepackage{orcidlink}
\newcommand{\vecmu}{\bm\mu}

\newcommand{\vecx}{\mathbf{x}}

\newcommand{\vecz}{\mathbf{z}}

\newcommand{\vectheta}{\bm\theta}

\newcommand{\vecvartheta}{\boldsymbol\vartheta}
\newcommand{\vecSigma}{\bm\Sigma}

\newtheorem{assumption}{Assumption}
\newtheorem{lemma}{Lemma}
\DeclarePairedDelimiter\abss{\lvert}{\rvert}

\newtheorem{definition}{Definition} 
 
\newtheorem{theorem}{Theorem}

\newcommand{\argmax}{\arg\max}

\newcommand{\bmX}{\bm{X}}

\newcommand{\MM}{\bm{M}}
\newcommand{\MV}{\bm{V}}
\newcommand{\MU}{\bm{U}}

\title{Clustering Three-Way Data with Outliers}
\author[1]{Katharine M. Clark\orcidlink{0000-0002-6162-2300}}
\author[2]{Paul D. McNicholas\orcidlink{0000-0002-2482-523X}}

\affil[1]{\small Department of Mathematics \& Statistics, Trent University, Ontario, Canada.}
\affil[2]{\small Department of Mathematics \& Statistics, McMaster University, Ontario, Canada.}

\date{} 
\begin{document}

\maketitle

\begin{abstract}
Matrix-variate distributions are a relatively recent addition to the model-based clustering literature, thereby making it possible to analyze data in matrix form with complex structure such as images and time series. Due to its recent appearance, there is limited literature on matrix-variate data, with even less on dealing with outliers in these models. An approach for clustering matrix-variate normal data with outliers is discussed. The approach, which uses the distribution of subset log-likelihoods, extends the OCLUST algorithm to matrix-variate normal data and uses an iterative approach to detect and trim outliers.\\
	
	\noindent\textbf{Keywords}: OCLUST, clustering, matrix-variate, mixture models, outliers, three-way data.

\end{abstract}
%
%
\section{Introduction}\label{sec:intro}
Matrix-variate normal mixture models are a powerful statistical tool used to represent complex data structures that involve matrices, such as multivariate time series, spatial data, and image data. Detecting outliers in matrix-variate normal mixture models is crucial for identifying anomalous observations that deviate significantly from the underlying distribution. Outliers can provide valuable insights into data quality issues, anomalies, or unexpected patterns. Outliers, and their treatment, is a long-studied topic in the field of applied statistics. The problem of handling outliers in multivariate clustering has been studied in several contexts including work by \cite{garcia08}, \cite{punzo16b}, \cite{punzo20}, and \cite{clark24}.

The OCLUST algorithm introduced in \cite{clark24}, and supported by the {\tt oclust} package \citep{oclust1.0.0} for {\sf R} \citep{R4.5.0}, is based on the mixture model-based clustering framework \citep[see, e.g.,][]{mcnicholas16b} and uses an iterative subset log-likelihood approach to detect and trim outliers. An analogue of the OCLUST algorithm is developed herein for three-way data.

The paper is organized as follows. In Section~\ref{sec:background}, matrix-variate normal mixtures and prior work to address outliers are reviewed. Then, the matOCLUST algorithm for clustering matrix-variate normal mixtures with outliers is developed (Section~\ref{sec:meth}) and illustrated via real and simulated data (Sections~\ref{sec:sim} and~\ref{sec: realdata}). The paper concludes with a discussion and some suggestions for future work (Section~\ref{sec:con}). 

\section{Background}\label{sec:background} 

\subsection{Matrix-Variate Normal Mixtures}
Classification creates a partition of a dataset to expose an underlying group structure present. The idea is to identify clusters, where data within a cluster are similar to each other, but dissimilar to data in other clusters. Clustering is a special case of classification, where cluster membership is unknown \textit{a priori}. 

Clustering algorithms generally fall into two categories: non-parametric and parametric methods. Non-parametric methods (e.g., $k$-means, hierarchical clustering) are usually distance-based, aiming to minimize the distance between points within a cluster and maximize distance to points in other clusters.

Alternatively, parametric methods, such as finite mixture models, aim to fit a model to the data. Each cluster is usually taken to correspond to a component with a specific probability density function. The density of a finite mixture model is
\begin{equation}
	f(\vecx\mid\vecvartheta)=\sum_{g=1}^{G}\pi_g f_g(\vecx\mid\vectheta_g),
	\label{eq:generalmixture1}
\end{equation}
where $\vecvartheta=\{\pi_1, \dots, \pi_G, \vectheta_1, \dots \vectheta_G\}$, $\pi_g>0$ is the $g$th mixing proportion with $\sum_{g=1}^G \pi_g =1$, and $f_g(\vecx\mid\vectheta_g)$ is the $g$th component density with parameters $\vectheta_g$. The likelihood can be maximized to to estimate the model parameters and, in turn, to determine cluster memberships. 

Traditionally, the component densities in \eqref{eq:generalmixture1} were chosen to Gaussian, but now mixture models employ a range of distributions, including multivariate skewed distributions \citep[e.g.,][]{franczak14,murray14b,dang23,mclachlin24,sochaniwsky25,pocuca26}, discrete distributions \citep[e.g.,][]{bouguila09,subedi20,silva23,payne25}, and matrix-variate distributions \citep[e.g.,][]{viroli11,gallaugher17,gallaugher18}. 

In this ``big-data" era,  data are becoming increasingly complex in terms of size and structure. While there is an extensive literature covering univariate and multivariate datasets, there is a paucity of work on higher-order data such as three-way, or matrix-variate, data. In a matrix-variate dataset, $N$ matrices are observed, e.g., in multivariate longitudinal data we observe $p$ quantities over $n$ time points for $N$ subjects and so we have $N$ $n\times p$ matrices. 


Three-way data can be clustered via matrix-variate mixture models, e.g., \cite{viroli11} uses a mixture of matrix-variate normal distributions. An $n\times p$ random matrix $\bmX$ comes from a matrix-variate normal distribution if its density is of the form
\begin{equation}\begin{split}
		\phi_{n\times p}({\bmX}~|~\MM, \MV, \MU ) = \frac{1}{(2\pi)^{ \frac{np}{2} } |\MV|^{\frac{n}{2}} |\MU|^{\frac{p}{2}}}\exp \left\{- \frac{1}{2}\text{tr}\big[\MV^{-1}(\bmX-\MM)^{'}\MU^{-1}(\bmX-\MM)\big]\right\}, 
		\label{eq:pdf}
\end{split}\end{equation}
where $\MM$ is the $n \times p$ mean matrix, $\MU$ is the $n \times n$ row covariance matrix, and $\MV$ is the $p \times p$ column covariance matrix.  
Note that the matrix-variate normal distribution is related to the multivariate normal distribution via the equivalence
\begin{equation}
	\mathscr{X}\sim \mathcal{N}_{n\times p}(\MM, \MU, \MV) \iff \text{vec}(\mathscr{X})\sim \mathcal{N}_{np}(\text{vec}(\MM), \MV \otimes \MU)
	\label{eq:norm}
\end{equation}
\citep{gupta99}, where $\otimes$ denotes the Kronecker product and $\text{vec}(\cdot)$ is the vectorization operator.
Note that there is an identifiability issue with regard to the parameters $\MU$ and $\MV$, i.e., if $a$ is a strictly positive constant, then 
\begin{equation*}
	\frac{1}{a}\MV\otimes {a} \MU=\MV\otimes\MU
\end{equation*}
and so 
replacing $\MU$ and $\MV$ by $(1/a)\MU$ and $a\MV$, respectively, leaves \eqref{eq:pdf} unchanged. With different restrictions, we can resolve this identifiability issue, such as setting $\text{tr}(\MU)=r$ or $\MU_{11}=1$ \citep[see][]{anderlucci15,gallaugher18}.

%

\subsection{Outlier Detection in Model-Based Clustering}
Failure to account for outliers in model-based clustering can have several implications. One issue is that the clustering results may be biased towards the outliers. Outliers can influence the estimation of the model parameters, which in turn can affect the clustering results. In addition, outliers, and their effect on the estimated model parameters, may make the clustering results less interpretable. When forced to be accommodated into the model, the outliers may move the component means or inflate their variances. This may lead one to draw inaccurate conclusions about the properties of each cluster. 

In model-based clustering, there are three main approaches to handling outliers: outlier-inclusion, outlier identification, and outlier trimming.
Outlier-inclusion methods, such as the one proposed by \cite{banfield93}, include outliers in an additional uniform component over the convex hull. This approach can be effective if the outliers are not cluster-specific and appear more like noise. Alternatively, if the outliers are cluster-specific and symmetric about the component-wise means, they can be accommodated by clustering with heavy-tailed distributions \citep[e.g.,][]{peel00, sun10, dang15,  bagnato17, tomarchio22b}. This concept has been extended to matrix-variate mixtures by \cite{tomarchio20}. Finally, outliers can be included in the model through the use of contaminated distributions. For multivariate Gaussian clusters with outliers, \cite{punzo16b} separate each cluster into two components --- the `good' points with density $\phi(\vecx\mid\vecmu_g,\vecSigma_g)$, and the `bad' points with density $\phi(\vecx\mid\vecmu_g,\eta_g\vecSigma_g)$. Each component has the same centre, but the `bad' points have a larger variance, where $\eta_g >1$. The principle of clustering with contaminated distributions has been extended to other multivariate distributions \citep[e.g.,][]{morris19, melnykov21, naderi24} and to the matrix-variate normal distribution by \cite{tomarchio22a}.

Outlier identification methods, such as the one proposed by \cite{evans15}, identify outliers after the model has been fit. Outlyingness of a point is determined by the degree the variance changes in its absence. This approach can be useful for novelty detection, as it can identify points that are significantly different from the rest of the data. However, it can be difficult to choose the appropriate threshold for identifying outliers.

Outlier trimming methods, such as the one proposed by \cite{cuesta97} and later modified by \cite{garcia08}, remove outliers from the dataset entirely. While based in the $k$-means domain, the TCLUST algorithm \citep{garcia08} allows for varying cluster weights and shapes by placing certain restrictions. However, required in advance, it can be difficult to choose the appropriate number of outliers to remove.

\subsection{OCLUST Algorithm}
The OCLUST algorithm \citep{clark24} is a trimming algorithm based in the model-based clustering domain. A subset log-likelihood is considered to be the log-likelihood of a model fitted with $N-1$ of the data points. The OCLUST algorithm uses the subset log-likelihoods and their distribution to identify and trim outliers.  The point $\vecx_k$, whose absence produced the largest subset log-likelihood, is treated as a candidate outlier. OCLUST continues removing candidate outliers until the `best' model is obtained, based on the distribution of the subset log-likelihoods, which is a mixture of shifted and scaled beta distributions \citep[see][for details]{clark24}.
This paper extends the OCLUST algorithm to matrix-variate normal data by deriving a similar distribution for the log-likelihoods of the subset models. 

\section{Methodology}\label{sec:meth}

\subsection{Distribution of Subset Complete-Data Log-Likelihoods}\label{sec:dist}

Consider the complete-data, which comprises the observed data $\bmX_1,\dots,\bmX_N$ and their cluster memberships $\vecz_1,\ldots,\vecz_N$, where $\vecz_i=(z_{i1},\ldots,z_{iG})'$, $z_{ig}=1$ if $\bmX_i$ belongs to the $g$th cluster, and $z_{ig}=0$ otherwise. The complete-data log-likelihood is given by
\begin{equation}\label{eq:completemat}
	l_{\mathcal{X}}=\sum_{i=1}^N \sum_{g=1}^G z_{ig}\left[\log\pi_g + \log\phi_{n \times p}(\bmX_i\mid \MM_g, \MU_g,\MV_g \right].
\end{equation}
As in the multivariate case, consider a subset complete-data log-likelihood in the matrix-variate case to be the compete-data log-likelihood of a model fitted with $N-1$ of the observation matrices. Formally, let the dataset be denoted by $\mathcal{X}=\{\bmX_1,\dots,\bmX_N\}$. The $j$th subset is defined as the dataset with the $j$th observation removed, i.e., $$\mathcal{X}\setminus \bmX_j=\{\bmX,\dots,\bmX_{j-1},\bmX_{j+1},\dots, \bmX_N\}.$$
We now introduce our main result.
\begin{theorem}\label{propmat}
	For a matrix $\bmX_j$ generated from a mixture of matrix-variate normal distributions, if $\bmX_j$ belongs to the $h$th cluster (i.e., $z_{jh}=1$),  $l_{\mathcal{X}}$ is the complete-data log-likelihood and $D_j=l_{\mathcal{X} \setminus \bmX_j}-l_{\mathcal{X}}$, then $D_j \mid (z_{jh}=1)$ has a shifted gamma density 
	\begin{equation}
		\label{eq:LM}
		D_j \mid (z_{jh}=1) \sim f_{\text{gamma}}\left( d_j-c_h~\bigg|~\frac{np}{2},1\right),
	\end{equation}
	for $d_j-c_h\geq0, \frac{np}{2}>0$, where $c_h=-\log\pi_h+\frac{np}{2}\log(2\pi)+\frac{c}{2}\log\abss{\MU_h}+\frac{r}{2}\log\abss{\MV_h},$ $N_h$ is the number of matrices in cluster h, and  $\pi_h=N_h/N$. 
\end{theorem}

\begin{proof}

	The quantity $l_{\mathcal{X}}$ can be regarded as the complete-data log-likelihood for the entire dataset~$\mathcal{X}$. Define $l_{\mathcal{X}\setminus \bmX_j}$ as the complete-data log-likelihood for the $j${th} subset $\mathcal{X}\setminus \bmX_j$. 
	
	Because $\pi_g$, $\MM_g$, $\MU_g$, and $\MV_g$, for $g=1,\ldots,G$, are population parameters, they are impervious to the sample drawn from the dataset and remain unchanged for each subset $\mathcal{X}\setminus \bmX_j$, for $j= 1,\ldots,N$. Thus, the complete-data log-likelihood for the $j$th subset, $\mathcal{X}\setminus \bmX_j$,  when $z_{jh}=1$ is 
	\begin{equation*}
		l_{\mathcal{X} \setminus \bmX_j} \mid (z_{jh}=1)=l_{\mathcal{X}}-\log{\pi_h}- \log\phi_{n \times p}(\bmX_j\mid \MM_g, \MU_g,\MV_g ).
	\end{equation*}
	Rearranging yields
	\begin{equation}\label{eq:likdiff}
		l_{\mathcal{X} \setminus \bmX_j}-l_{\mathcal{X}}\mid (z_{jh}=1)=-\log\pi_h+\frac{rc}{2}\log(2\pi)+\frac{c}{2}\log\abss{\MU_h}+\frac{r}{2}\log\abss{\MV_h} + \frac{1}{2}T_j, 
	\end{equation}
	where $T_j=\text{tr}\big(\MV_h^{-1}(\bmX_j-\MM_h)^{'}\MU_h^{-1}(\bmX_j-\MM_h) \big)$  is the matrix-variate Mahalanobis distance. Because $ T_j \sim \chi^2_{np} = \text{gamma}\left(\frac{np}{2},2\right) $ \citep{gupta99}, the scaling property of the gamma distribution implies that $ \frac{1}{2}T_j \sim \text{gamma}\left(\frac{np}{2},1\right). $ Let $ c_h=-\log\pi_h+\frac{np}{2}\log(2\pi) +\frac{c}{2}\log|\MU_h| +\frac{r}{2}\log|\MV_h|, $ and note that $ D_j=c_h+\frac{1}{2}T_j.$ Therefore, $ D_j-c_h \sim \text{gamma}\left(\frac{np}{2},1\right), $ and 
 \begin{equation}
  f_{D_j\mid (z_{jh}=1)}(d_j) = f_{\text{gamma}}\left( d_j-c_h \,\bigg|\, \frac{np}{2},1 \right), \qquad d_j\ge c_h. 
  \label{eq:ydensmat}
 \end{equation}
\end{proof}

Let \eqref{eq:ydensmat} represent the conditional density for $D_j$ when $z_{jh}=1$. Denote this density as $f_h(d)$. The density for $D$ unconditional on the cluster membership is then
\begin{equation}\label{eq:ymix}
	f_D(d|\vecvartheta)=\sum_{g=1}^G \pi_g f_g(d|\vectheta_g),
\end{equation}
where $f_g(d|\vectheta_g)$ has the shifted gamma density described in \eqref{eq:ydensmat}, and $\vectheta_g=(\pi_g,n,p,\MU_g,\MV_g)$

\subsection{MatOCLUST Algorithm}\label{sec:oclust}
Thus far, $l$ has represented the complete-data log-likelihood. Now let a `hat' denote this quantity calculated with the maximum likelihood estimates (MLEs), i.e. 
\begin{itemize}
	\item[$l$] denotes the complete-data log-likelihood in \eqref{eq:completemat} with population parameters $\pi_g$, $\MM_g$, $\MU_g$, and $\MV_g$ for $g=1,\ldots,G$; and 
	\item[$\hat l$] denotes the complete-data log-likelihood in \eqref{eq:completemat} with MLEs $\hat\pi_g$, $\hat\MM_g$, $\hat\MU_g$, and $\hat\MV_g$ for $g=1,\ldots,G$. 
\end{itemize}
Now consider the observed log-likelihood for a matrix-variate normal mixture model
\begin{equation}
	\ell_{\mathcal{X}}=\sum_{i=1}^N \log \left[\sum_{g=1}^G \pi_g \phi_{n \times p}(\bmX_i\mid \MM_g, \MU_g,\MV_g )\right].
	\label{eq:loglikmat}
\end{equation}
Similarly, let a `hat'  denote the estimated version, where
\begin{itemize}
	\item[$\ell$] denotes the observed log-likelihood in \eqref{eq:loglikmat} with population parameters $\pi_g$, $\MM_g$, $\MU_g$, and $\MV_g$ for $g=1,\ldots,G$;
	\item[$\hat\ell$] denotes the observed log-likelihood in \eqref{eq:loglikmat} with MLEs $\hat\pi_g$, $\hat\MM_g$, $\hat\MU_g$, and $\hat\MV_g$ for $g=1,\ldots,G$. 
	\end{itemize}

The matOCLUST algorithm proceeds by removing candidate outliers one-by-one until the best model is achieved, which is determined by the distribution of the subset complete-data log-likelihoods, stated in Theorem~\ref{propmat}. Analogous to the multivariate case, treat matrix $\bmX_k$, whose absence produced the largest subset log-likelihood, as the candidate outlier; see Definition~\ref{def:outmat}. 
\begin{definition}[Candidate Outlier] \label{def:outmat} Define the candidate outlier as $\bmX_k$, where
	\begin{equation*}
		k=\argmax_{j \in [1,N]} \hat\ell_{\mathcal{X} \setminus \bmX_{j}}, 
		\label{eq:argmaxmat}
	\end{equation*}
	and $\hat\ell_{\mathcal{X} \setminus \bmX_{j}}$ is the estimated observed log-likelihood of the subset model with the $j$th observation removed. \label{def:matout}\end{definition}
Note that this definition uses the observed log-likelihood estimated with MLEs, which represents a  change from the complete-data log-likelihood used in Section~\ref{sec:dist}. This choice is practical because, in practice, the cluster memberships (and therefore the complete-data) are unknown.  This requires the following assumption and lemma. 
\begin{assumption}\label{ass:sep}
	The clusters are non-overlapping and well separated.
\end{assumption}
This assumption supports the conditions necessary for Lemma~\ref{lem:loglikmat}. 
\begin{lemma}\label{lem:loglikmat}
	As cluster separation tends to infinity, $\ell_{\mathcal{X}} \rightarrow l_{\mathcal{X}}$.  In other words, the complete-data log-likelihood is the limiting form of the observed log-likelihood.
\end{lemma}
\begin{proof} \cite{clark24} prove the analogous result for multivariate normal random variables and the same property applies to matrix-variate normal random variables through the equivalence 
	\begin{equation*}
		\mathscr{X}\sim N_{n \times p}(\MM, \MU, \MV) \iff \text{vec}(\mathscr{X})\sim N_{np}(\text{vec}(\MM), \MV \otimes \MU).
	\end{equation*}
\end{proof}

Lemma~\ref{lem:loglikmat} allows the use of ${\ell}_{\mathcal{X}}$  and ${\ell}_{\mathcal{X} \setminus \bmX_j}$ in place of ${l}_{\mathcal{X}}$ and ${l}_{\mathcal{X} \setminus \bmX_j}$, respectively, when the clusters are sufficiently separated. Furthermore, because the true parameters $\pi_g$, $\bm{M}_g$, $\bm{U}_g$,  and $\bm{V}_g$ are typically unknown in the clustering paradigm, we must use $\hat{\ell}_{\mathcal{X} \setminus \bmX_j}$ and $\hat{\ell}_{\mathcal{X}}$,  the observed full and $j$th subset model log-likelihoods, respectively, from \eqref{eq:loglikmat} using the MLEs. The result is an approximation for \eqref{eq:LM}, formally given in Theorem~\ref{theo:est}, 

\begin{theorem}\label{theo:est}
Let $\hat D_j =\hat{\ell}_{\mathcal{X} \setminus \bmX_j} -\hat{\ell}_{\mathcal{X}}$, then 
\begin{equation}\label{eq:approxdist}
	\hat D_j \mid( z_{jh}=1) \hspace{5pt} \dot{\sim} \hspace{5pt}f_{\text{gamma}}\left( \hat{d}_j-\hat{c}_h~\bigg|~\frac{np}{2},1\right),
\end{equation}
where $$\hat{c}_h=-\log\hat{\pi}_h+\frac{np}{2}\log(2\pi)+\frac{c}{2}\log\abss{\hat{\MU}_h}+\frac{r}{2}\log\abss{\hat{\MV}_h}.$$ 
\end{theorem}

\begin{proof}
	A proof is given in Appendix~\ref{app:approx}.
\end{proof}

Result \eqref{eq:approxdist} describes the distribution of the difference between the subset observed log-likelihood and the observed log-likelihood for the full model when $z_{jh}=1$. Just like in \eqref{eq:ymix}, the density for $\hat D_j $ unconditional on the cluster membership of $\bmX_j$ is given by the following mixture model:
\begin{equation}\label{eq:approxmixy}
	f(\hat{d}|\vecvartheta)=\sum_{g=1}^G \hat\pi_g f_g(\hat{d}|\hat\vectheta_g),
\end{equation}
where $f_g(\hat{d}|\hat\vectheta_g)$ has the shifted gamma density described in \eqref{eq:approxdist} and $\hat\vectheta_g=(\hat\pi_g,n,p,\hat\MU_g,\hat\MV_g)$.

Equation \eqref{eq:approxmixy} is the density of the approximate null distribution for the difference in observed log-likelihoods between the subset and full models when the data arise from a mixture of matrix-variate normal distributions. Thus, divergence from this distribution indicates that the clusters are not strictly matrix-variate normal. While this could indicate that the data arise from another distribution entirely, we will assume that the choice of a matrix-variate normal model is correct and that the presence of outliers is the cause of this discrepancy. 

This null distribution is leveraged to create an outlier trimming algorithm. Most-likely outliers are removed iteratively and the Kullback-Leibler (KL) divergence from the null distribution is calculated. Outliers continue to be removed one-by-one until a pre-specified number is achieved.  The chosen model is the one that minimizes the KL divergence from the null gamma distribution. With each iteration, the candidate outlier is chosen according to Definition~\ref{def:matout}. The result is the matOCLUST algorithm, described in Algorithm~\ref{matOCLUST}. Note that $F$ is a chosen upper bound that corresponds to the maximum number of candidate outliers to be considered. 

\begin{algorithm}[!htb]
	\caption{matOCLUST algorithm}\label{matOCLUST}
	\begin{algorithmic}[1]
		\Procedure{matOCLUST}{$\mathcal{X},G,F$}
		\State (Optional) Identify gross outliers, $b$, using method of choice. $B=\#b$. 
		\State \begin{varwidth}[t]{\linewidth}
			Update:\par
			\hskip\algorithmicindent 	$\mathcal{X} \hookleftarrow \mathcal{X} \setminus b$ \par 
			\hskip\algorithmicindent 	$N \hookleftarrow N-B$ 
		\end{varwidth}\newline
		\For{$f $ in $B:F$}
		\State \begin{varwidth}[t]{\linewidth}
			Cluster the data $\mathcal{X}$ into $G$ clusters, using a matrix-variate normal model- \newline
			\hskip\algorithmicindent based clustering algorithm \label{step: clustermat}	\end{varwidth} \newline
		\State Output $\hat\ell_{\mathcal{X}}$, and for each cluster: $\hat\MM_g$, $\hat\MU_g$, $\hat\MV_g$, $\hat\pi_g=\hat N_g/N$. \label{step:parasmat}
		\For{$j$ in $1:N$}
		\State \begin{varwidth}[t]{\linewidth}
			Cluster the subset $\mathcal{X} \setminus \bmX_j$ into $G$ clusters, using the chosen method\newline
			\hskip\algorithmicindent in Step \ref{step: clustermat}.	\end{varwidth} \newline
		\State Output $\hat\ell_{\mathcal{X} \setminus \bmX_j}$ and calculate $\hat{d}_j=\hat\ell_{\mathcal{X} \setminus \bmX_j}-\hat\ell_{\mathcal{X}}$.
		\EndFor
		\State \label{step:densmat}\begin{varwidth}[t]{\linewidth} Generate the density of $\hat D$ using \eqref{eq:approxmixy} and the parameters from \newline
			\hskip\algorithmicindent Step~\ref{step:parasmat}.
		\end{varwidth} \newline
		\State \begin{varwidth}[t]{\linewidth}
			Calculate the approximate KL divergence of $\hat{d}_1, \dots, \hat{d}_N$ from the density\newline
			\hskip\algorithmicindent in Step~\ref{step:densmat} using relative frequencies.	\end{varwidth} \newline
		
		\State Determine the most likely outlier $\bmX_k$ as per Definition~\ref{def:outmat}.
		\State \begin{varwidth}[t]{\linewidth}
			Update:\par
			\hskip\algorithmicindent 	$\mathcal{X} \hookleftarrow \mathcal{X} \setminus \bmX_k$ \par
			\hskip\algorithmicindent 	$N \hookleftarrow N-1$
		\end{varwidth}\newline
		\EndFor
		\State Choose $f$ for which KL divergence is minimized \newline 
		\Comment{This is the predicted number of outliers.} \newline
		\Comment{Use the model corresponding to iteration $f$.}
		\EndProcedure
	\end{algorithmic}
\end{algorithm}

\section{Simulation Study}\label{sec:sim}
This section evaluates matOCLUST's performance against six of its competitors: three matrix-variate clustering models (a.--c.) and three outlier-specific multivariate normal clustering algorithms (d.--f.).
\begin{enumerate}[label=\alph*.]
	\item Matrix-variate contaminated normal mixtures \citep[MVCNM;][]{tomarchio22a};  
	\item Matrix-variate normal mixtures \citep[MVN;][]{viroli11};
	\item Mixtures of matrix-variate t-distributions \citep[MVT;][]{dougru16}
	\item The (multivariate) OCLUST algorithm \citep{clark24};  
	\item (Multivariate) contaminated normal mixtures \citep[CNMixt;][]{punzo16b};  
	\item (Multivariate) noise component mixtures, mixtures of Gaussian clusters and a uniform component \citep[NCM;][]{banfield93}. 
\end{enumerate}
\vspace{5pt}
For each of the multivariate methods, the data are first vectorized. Each algorithm is applied to three types of dataset.

\subsection{Simulated Datasets}\label{sec:data}

\subsubsection{In the Style of \cite{tomarchio22a}}\label{sec:tomarchio}

The first type closely follows the scheme in \cite{tomarchio22a}. A total of 100 datasets are created with $N=200$ $2\times 4$ matrices in two groups of equal weight with parameters
\begin{equation*}
	\MM_1=
	\begin{pmatrix}
		-2.60 & -1.10 & -0.50 & -0.20\\
		1.30 & 0.60 & 0.30 & 0.10
	\end{pmatrix},\hspace{10pt}
	\MM_2=
	\begin{pmatrix}
		1.50 & 1.70 & 1.90 & 2.20\\
		-3.70 & -2.70 & -2.00 & -1.50
	\end{pmatrix},
\end{equation*}
\begin{equation*}
	\MU_1=
	\begin{pmatrix}
		2.00 & 0.00 \\
		0.00 & 1.00
	\end{pmatrix},\hspace{10pt}
	\MU_2=
	\begin{pmatrix}
		1.70 & 0.50 \\
		0.50 & 1.30
	\end{pmatrix},
\end{equation*}
\begin{equation*}
	\MV_1=\MV_2=
	\begin{pmatrix}
		1.00 & 0.50 & 0.25 & 0.13\\
		0.50 & 1.00& 0.50 &0.25\\
		0.25 &0.50 &1.00 &0.50\\
		0.13 &0.25 &0.50 &1.00
	\end{pmatrix}.
\end{equation*}
Ten matrices are randomly selected, and then one column from each selected matrix is randomly replaced with data generated from $\text{Uniform}(-15,15)$.

\subsubsection{In the Style of \cite{viroli11}}\label{sec:viroli}
The second type generates data using the scheme in \cite{viroli11}. A total of 100 datasets are created with $N=300$ $3 \times 5$ matrices in three groups with proportions $(\pi_1, \pi_2, \pi_3) = (0.3, 0.4, 0.3)$ and means of the form
\begin{equation*}
	\MM_1=
	\begin{pmatrix}
		0.5 & 0& 0& 0& 0\\
		0.5 & 0& 0& 0& 0\\
		0 & 0& 0& 0& 0
	\end{pmatrix},\hspace{10pt}
	\MM_2=
	\begin{pmatrix}
		0 & 0& 0& 0& 0\\
		0 & 0& 0& 0& 0\\
		0 & 0& 0& 0& 0
	\end{pmatrix}, \hspace{10pt}
	\MM_3=
	\begin{pmatrix}
	-0.5& 0& 0& 0& 0\\
	0.5 & 0& 0& 0& 0\\
	0 & 0& 0& 0& 0
	\end{pmatrix}.
\end{equation*}
The matrices $\MU_g$ and $\MV_g$ of dimensions $3\times 3$ and $5 \times 5$, respectively, are randomly generated using the \texttt{rcorrmatrix} function in the \texttt{clusterGeneration} \citep{clustergeneration} package for {\sf R} \citep{R4.5.0}, which uses the methodology in \cite{joe06}. 

Where the simulations differ is in the generation of outliers. To simulate potential outliers due to errors in data-entry, 15 matrices are randomly selected and the entries in each matrix are randomly permuted. 

\subsubsection{Clean}\label{sec:clean}
Finally, to assess the specificity of matOCLUST, it and its competitors are applied to the Viroli-style datasets in Section~\ref{sec:viroli} without generated outliers. These are the exact same datasets generated before the entries were permuted.

\subsection{Results}

\begin{figure}[htb]
	\centering
	\includegraphics[width=\textwidth]{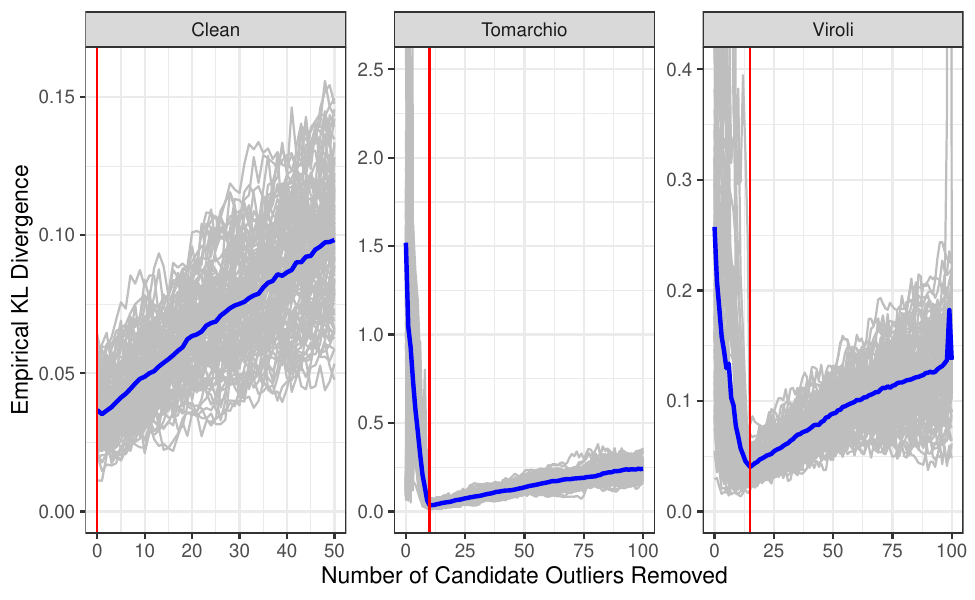}
		\caption{ Plots of KL calculated at each iteration for 100 datasets of each type.  The blue curve overlaid represents the mean of the 100 runs and red vertical line indicates the number of simulated outliers.}
	\label{fig:KL}
\end{figure}

The MVCNM, MVN, and MVT methods were run in \textsf{R} using the \texttt{MatrixMixtures} \citep{MatrixMixtures} package. The OCLUST, CNMixt, and NCM methods were implemented in the \texttt{oclust} \citep{oclust1.0.0}, \texttt{ContaminatedMixt} \citep{punzo18}, and \texttt{mclust} \citep{mclust6.1.3} packages, respectively. The matOCLUST algorithm was implemented in Julia \citep{julia} on each type of dataset, setting the maximum number of outliers to $F=100$ for the Viroli-style and Tomarchio-style datasets and to $F=50$ for the clean datasets. The resulting KL divergence is plotted in Figure~\ref{fig:KL}, with the actual number of simulated outliers indicated as a red vertical line.  

\begin{figure}[!htb]
	\centering
	\includegraphics[width=\textwidth]{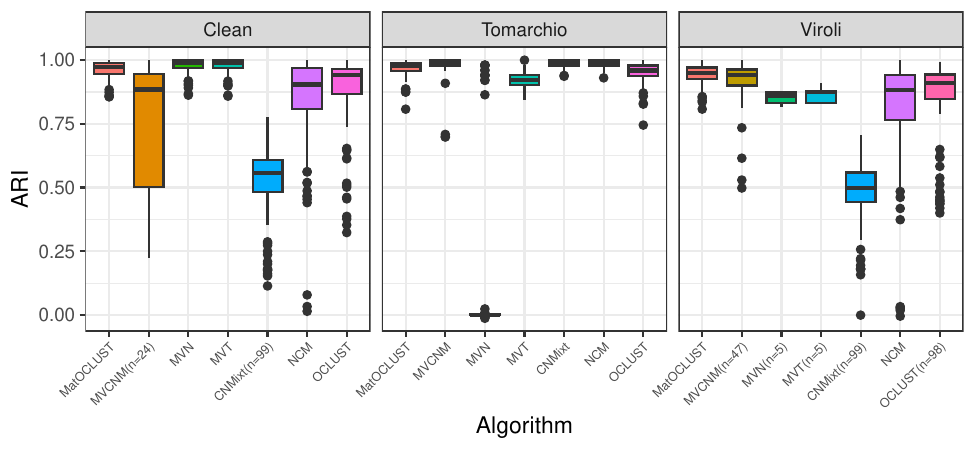}
	\caption{A boxplot of ARI for each algorithm on the simulated datasets.}
	\label{fig:ARI}
\end{figure}

\begin{figure}[!htb]
	\centering
	\includegraphics[width=\textwidth]{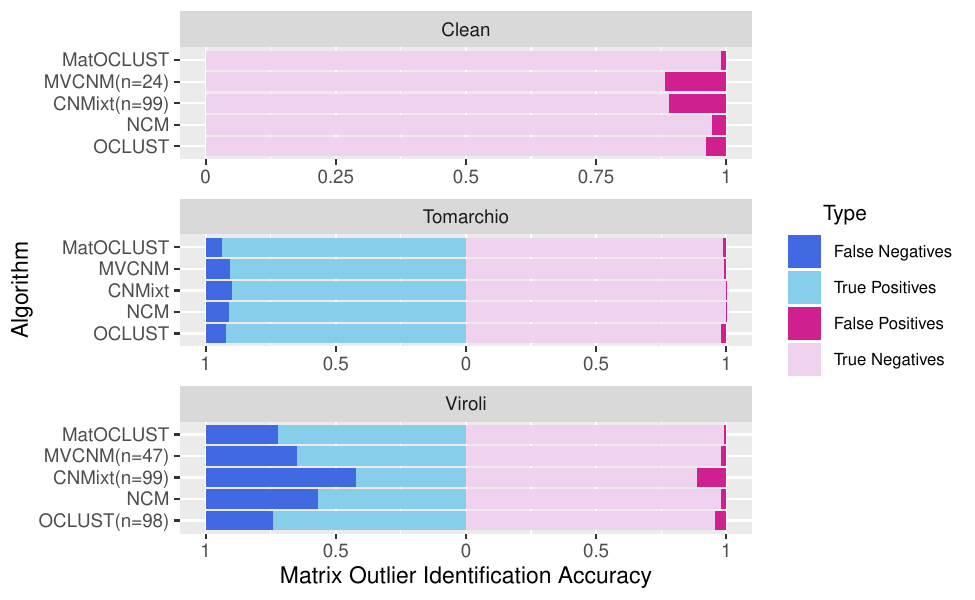}
	\caption{Outlier classification rates for each algorithm on the three styles of datasets. }
	\label{fig:errors}
\end{figure}

Performance for each algorithm in terms of adjusted Rand index \citep[ARI;][]{hubert85} is displayed in Table~\ref{tab:ARI} and Figure~\ref{fig:ARI}, while false positive, true negative, true positive and false negative rates are plotted in Figure~\ref{fig:errors}. Note that for ARI, methods that identify outliers are evaluated on a solution with `outlier' as its own class. For MVN and MVT, ARI is calculated considering the outlier's class before permutation or contamination. 
\begin{table}[!htb]
	\centering
	\caption{Mean ARI for each algorithm on the simulated datasets, with standard deviation in parentheses and number of successful runs in square brackets.}
	\begin{tabular}{lcccccc}
		\hline 
		&& Clean && Tomarchio &&  Viroli\\
		\hline
		
		matOCLUST && 0.962 (0.035) [100]&&0.967 (0.033) [100]&& \textbf{0.944} (0.039) [100]\\ 
		MVCNM && 0.752 (0.261) [24]&&  0.977 (0.051) [100]&&  0.905  (0.108) [47] \\ 
		MVN && 0.976 (0.030) [100]&&  0.133 (0.331) [100]&&  0.852 (0.026) [5] \\ 
		MVT && \textbf{0.977} (0.030) [100]&&  0.918  (0.031) [100]&&  0.864 (0.035) [5]\\ 
		CNMixt && 0.521 (0.143) [99]&&   0.984 (0.014) [100]&& 0.479 (0.129) [99]\\ 
		NCM && 0.835 (0.204) [100]&& \textbf{0.985 }(0.014) [100]&&0.799 (0.229) [100]\\ 
		OCLUST &&  0.873 (0.162) [100]& &0.951 (0.041) [100]&&0.846 (0.163)[98]\\
			\hline
		
	\end{tabular}\label{tab:ARI}
\end{table}

\subsubsection{Data in the Style of \cite{tomarchio22a}}

All algorithms perform very well with the exception of MVN, which has no robustness features. Matrix-variate and multivariate algorithms perform equally well, suggesting that the matrix-variate structure is not critical to discern the clusters. This likely stems from the mean matrices being so well separated. The approximately 10\% false negative error rate is consistent across algorithms and likely a result of the outlier generation process. By replacing a random column with random numbers, it is likely that some of the generated numbers fall within the range of typical observations, making the matrices appear inlying. 

\subsubsection{Data in the Style of  \cite{viroli11}}

These datasets were the most difficult to fit with the \texttt{MatrixMixtures} package, with only 47, five, and five successful runs from MVCNM, MVN, and MVT, respectively.  MatOCLUST and NCM were able to fit all datasets successfully, while CNMixt and OCLUST failed on one and two, respectively. Thus, the results for each algorithm only reflect the successful runs, and no valid conclusions can be drawn for MVN and MVT. 

All algorithms have a very high false negative rate, ranging from 26--58\%. However, due to the method of generation of outliers, it is very likely that some of the permuted matrices are not in any way unusual, given the structure of the data. The mean matrices are sparse, and the means of group 2 are zero in each dimension. Because the expectation of each permuted variable is the same in group 2, matrices from that group will only be unusual if entries with a large variance --- and large realizations of those variables --- are swapped with entries with low variance. For this reason, all algorithms have a large false negative rate. MatOCLUST has by far the lowest false positive rate, misidentifying approximately half as many as its next lowest competitor, MVCNM (1\% vs.\ 2\%).  MatOCLUST has the second lowest false negative rate. 

Overall, the matrix-variate methods perform better than the vectorized multivariate methods in terms of ARI. This likely comes down to the structure of the data. The mean matrices are very sparse, with separation only occurring in a very small two-dimensional subspace. This makes covariance estimation crucial, which only the matrix-variate methods are able to perform correctly. 

\subsubsection{Clean Data}

While identifying outliers is the goal for matOCLUST, it is arguably just as important, if not more so, to be able to distinguish when a dataset does not contain outliers. MatOCLUST, MVN and MVT all perform similarly well on the clean dataset, with matOCLUST achieving a  slightly lower ARI. This implies that these three algorithms all have a good ability to not find outliers if they are not present. All remaining algorithms perform noticeably worse, with lower ARIs and higher false positive rates. MVCNM's high false positive rate likely contributes to its low ARI because there are many observations classified into a fourth `outlier' class that does not exist in the original data. 

As seen in the Viroli-style datasets, the multivariate methods perform considerably worse than the matrix-variate versions, indicating that the matrix-variate structure is important to discern the clusters. MatOCLUST has the lowest false positive rate among its competitors at less than half that of NCM, showing good specificity. Overall, matOCLUST performs well in all three scenarios, showing strong and consistent results. 

\section{Real Data} \label{sec: realdata}
Finally, the matOCLUST algorithm is applied to two datasets to test its applicability on real data. 

\subsection{ANVUR Dataset} \label{sec:ANVUR}
The first example concerns data from the National Agency for the Evaluation of Universities and Research Institutes (ANVUR) on university programs in Italy. The data, first used by \cite{tomarchio22a}, consist of three metrics on first-year students taken from a three-year period. ANVUR records the percentage of credits earned by students out of the total required (Credits), the percentage of students that continue into their second year (Continue), and the percentage of students who would re-enroll  in the same program again (ReEnroll). The data are reported as averages across universities for each of 75 different degree families. Thus, there are 75 observations of  $3\times 3$ matrices. Of these degree families, 33 are undergraduate programs, while 42 are graduate programs, comprising our two groups. 

\begin{figure}[!htb]
	\centering
	\includegraphics[width=0.5\textwidth]{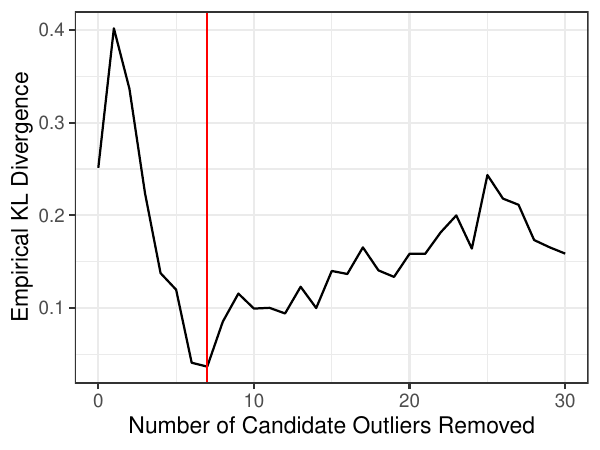}
	\caption{KL graph for the ANVUR dataset, where the red vertical line indicates the global minimum at seven candidate outliers.}
	\label{fig:ANVURKL}
\end{figure}

\begin{figure}[!htb]
	\centering
	\includegraphics[width=6in]{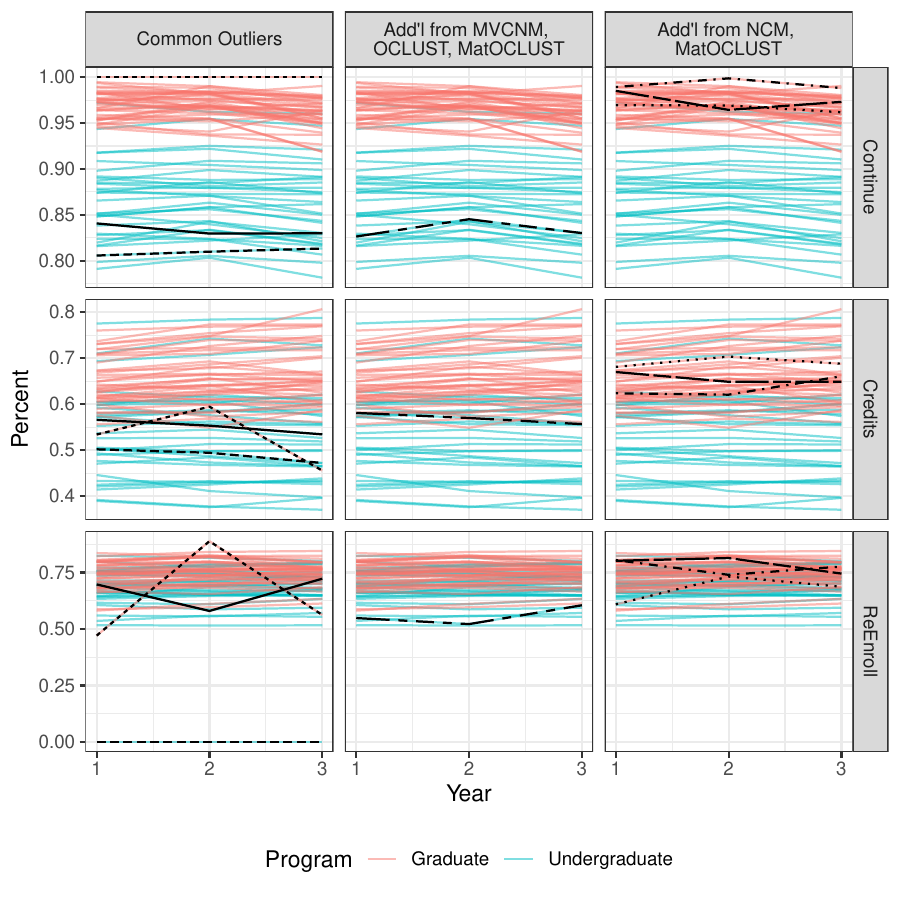}
	\caption{Line plots for the three variables (rows) across years ($x$-axis) for the ANVUR dataset. Lines are coloured by true class. In each panel, detected outliers are coloured black. Columns correspond to which algorithm detects the outlier.}		\label{fig:ANVUR}
\end{figure}

All of the algorithms from the simulation study are applied, pre-specifying $G=2$ for each. As seen in Figure~\ref{fig:ANVURKL}, minimum KL is achieved at 7 outliers, so the final model is from that iteration. A confusion matrix for each model and a line plot of the data can be seen in Table~\ref{tab:ANVUR} and Figure~\ref{fig:ANVUR}, respectively. All algorithms with the exception of MVN can discern the cluster structure, indicating that these data require a robust clustering method. 
\begin{table}[!htb]
	\centering
	\caption{Classification results for matOCLUST and MVCNM on the ANVUR dataset.}
	\begin{tabular}{@{\extracolsep{\fill}}llccccccccccccc}
		\hline 
		& &  \multicolumn{3}{c}{matOCLUST}&&\multicolumn{3}{c}{MVCNM}&&\multicolumn{2}{c}{MVN} &&\multicolumn{2}{c}{MVT} \\
		\cline{3-5}\cline{7-9}\cline{11-12}\cline{14-15}
		{Program Type}&&1&2&bad&&1&2&bad&&1&2&&1&2\\
		\hline 
		Undergraduate && 27 &3 &3 && 27& 3 &3&&30&3&&29&4\\ 
		Graduate& &  &38&4& &&41&1&&39&3&&&42\\ 
		&&&&&&&&&&&&&&\\
		&&\multicolumn{3}{c}{CNMixt} &&&\multicolumn{3}{c}{NCM} &&&\multicolumn{3}{c}{OCLUST}\\
		\cline{3-5}\cline{8-10}\cline{13-15}
		{Program Type}&&1&2&bad&&&1&2&bad&&&1&2&bad\\
		\hline 
		Undergraduate && 28 &3 &2 &&& 27& 4 &2&&&27&3&3\\ 
		Graduate &&  &41&1& &&&38&4&&&2&39&1\\ 
			\hline 
	\end{tabular}\label{tab:ANVUR}
\end{table}

The graphs in Figure~\ref{fig:ANVUR} give a better look at the structure of the data. All methods (with the exception of MVN and MVT which do not identify outliers) detect the same three outliers, as shown in the left column. Their anomalous behaviour is best seen in the ReEnroll variable in the bottom row. One program records zero in all years and the other two show swings in extremes, where a program among the lowest in re-enrolment in years~1 and~3 becomes one of the highest in year~2 and vice versa for the other curve. 

\begin{figure}[!htb]
	\centering
	\includegraphics[width=4.5in]{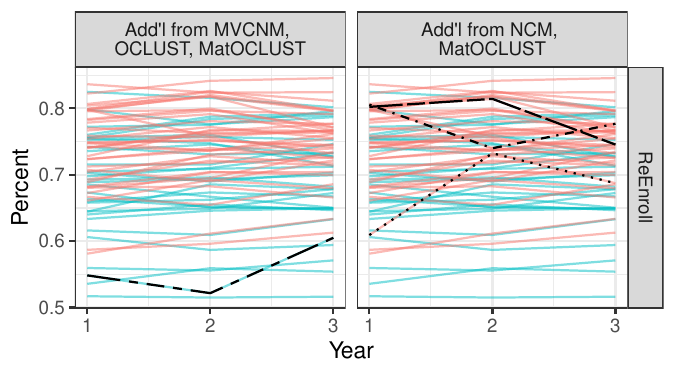}
	\caption{Centre and right panels for the ReEnroll variable of Figure~\ref{fig:ANVUR} magnified by excluding the extreme outlier. }	\label{fig:ANVURzoom}
\end{figure}

 The centre column shows the additional outlier detected by matOCLUST, OCLUST, and MVCNM. The rightmost column shows the additional outliers identified by NCM and matOCLUST. These outliers are also most evident in the ReEnroll variable. Their properties are best seen in Figure~\ref{fig:ANVURzoom}, where the anomalous curve with zero re-enrolment  is excluded. In the left panel, the curve is concave up where most curves with similar values are concave down. In the right panel, the lines are at a sharper angle, showing a larger discrepancy in year~2. These observations are more obviously outlying once the extreme outlier is removed --- a clear advantage of the iterative process matOCLUST uses for removing outliers. In total, there are seven unique outliers identified across competitor algorithms, which are all detected by matOCLUST. 

\subsection{LANDSAT Dataset} \label{sec:LANDSAT}

The second real dataset concerns the Landsat data from the UCI Machine Learning Repository \citep{landsat}.  Each observation represents the values from four spectral bands in a 3×3 neighbourhood of pixels from a Landsat satellite image. Thus, the observations can be expressed as $4\times 9$ matrices. Each pixel neighbourhood is classified into one of seven types of ground cover. This study is restricted to grey soil, damp grey soil, and soil with vegetation stubble soil types leaving $N=845$ matrices. 

\begin{figure}[!htb]
	\centering
	\includegraphics[width=0.5\textwidth]{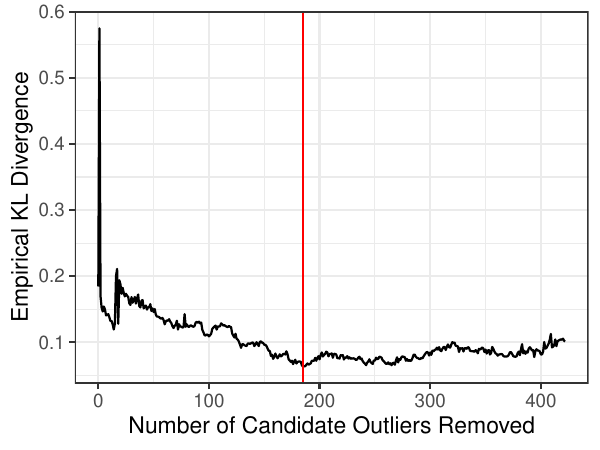}
	\caption{KL graph for the Landsat dataset. The vertical red line indicates the global minimum at 185 candidate outliers.}
	\label{fig:LANDSATKL}
\end{figure}

Each algorithm from Section~\ref{sec:sim} is run with $G=3$ clusters. The confusion matrices for the three best-performing methods are shown in Table~\ref{tab:LANDSAT}. MatOCLUST identifies 185 outliers where KL  is minimized, as seen in Figure~\ref{fig:LANDSATKL}.  Meanwhile, MVCNM identifies 218 pixel neighbourhoods as anomalous. Only matOCLUST and MVT are able to distinguish between grey soil and damp grey soil, with each misclassifying 62 and 111 images, respectively. To facilitate a more direct comparison with MVT, the observations identified as outliers by matOCLUST can be reclassified using the maximum posterior probabilities from the final fitted matOCLUST model. After reclassifying, there are 117 misclassified images. The confusion matrix for the reclassified points is shown in Table~\ref{tab:LANDSAT} and denoted by `matOCLUST-r'. 

The MVT and matOCLUST algorithms produce similar results with unique advantages. MVT delivers slightly superior performance when using all data. Instead, matOCLUST is uniquely able to identify spurious images. This may prove to be practical, such as creating a method for researchers to isolate regions to target for further analysis. While performance is comparable after the outliers are reclassified, matOCLUST particularly excels once the outlying observations are removed from the dataset.

\begin{table}[!htb]
	\centering
	\caption{Classification results for matOCLUST and MVCNM on the ANVUR dataset.}
	\begin{tabular}{@{\extracolsep{\fill}}llccccccccc}
		\hline 
		& &  \multicolumn{4}{c}{matOCLUST}&&\multicolumn{4}{c}{matOCLUST-r}\\
		\cline{3-6}\cline{8-11}
		{Soil Type}&&1&2&3&bad&&&1&2&3\\
		\hline 
		grey soil
		&& 303&37 & &57&&& 333& 56 &8\\ 
		damp grey soil
		&&  20&159&5&27& &&20&164&27\\ 
		soil with vegetation stubble &&  &&136&101& &&&6&231\\ 
		&&&&&&&&&&\\
			&&\multicolumn{4}{c}{MVCNM} &&&\multicolumn{3}{c}{MVT}\\
		\cline{3-6}\cline{9-11}
		{Soil Type}&&1&2&3&bad&&&1&2&3\\
		\hline 
		grey soil	&&304& 2 &&91&&&338&51&8\\ 
		damp grey soil	&&122&27&9&53&&&22&162&27\\ 
		soil with vegetation stubble &&&14&149&74&&&&3&234\\ 
		
		\hline 
	\end{tabular}\label{tab:LANDSAT}
\end{table}

\section{Conclusion}\label{sec:con}
The OCLUST algorithm has been extended to matrix-variate normal mixture models to facilitate clustering with outliers. The distribution of the difference between the log-likelihood of the subset model with one matrix removed and the full model has been shown to have an approximate shifted-gamma null distribution. The result is the matOCLUST algorithm, which removes candidate outliers iteratively until the difference in log-likelihoods no longer diverges from the null distribution. 
In simulations, matOCLUST is the only algorithm to perform well in all three considered scenarios. It has among the lowest false positive rates, indicating that it rarely unnecessarily removes `good' observations. In one real data application, matOCLUST identifies unusual yearly trajectories and, in another, it delivers superior classification performance once outliers are removed. 
Future work will be to extend the OCLUST algorithm to situations where clusters are not well captured via Gaussian distributions, e.g.,  to be able to identify outliers when there is significant skewness in the clusters.

\section*{Acknowledgements}
This work was funded by an NSERC Discovery Grant, an NSERC Canada Graduate Scholarship, a Dorothy Killam Fellowship, and the Canada Research Chairs program. 

\section*{Code Availability}
A software implementation of matOCLUST is available on GitHub \citep{clark26}.

\bibliographystyle{chicago}
\bibliography{masterbib.bib}

\appendix

\section{Proof of Theorem \ref{theo:est}} \label{app:approx} 

The aim is to show that, when the clusters are well separated ($s\to\infty$) and the sample size is large ($N\to\infty$), 
\begin{equation} 
	\hat{\ell}_{\mathcal X\setminus \bmX_j} - \hat{\ell}_{\mathcal X} \overset{P}{\longrightarrow} l_{\mathcal X\setminus \bmX_j} - l_{\mathcal X}.
 \end{equation} 
 Let 
\begin{equation} 
\boldsymbol\vartheta = \{\pi_g, \MM_g,\MU_g,\MV_g\}_{g=1}^{G}, 
\end{equation} 
and let 
\begin{equation} 
\hat{\boldsymbol\vartheta}_N = \{\hat\pi_g, \hat{\MM}_g,\hat{\MU}_g,\hat{\MV}_g\}_{g=1}^{G} 
\end{equation} 
denote the corresponding maximum likelihood estimators based on a sample of size $N$. Because the matrix-variate normal and multinomial distributions belong to the exponential family, the maximum likelihood estimators exist and are consistent \citep{gupta99,dasgupta08}. Hence, 
\begin{equation} 
	\hat{\boldsymbol\vartheta}_N \overset{P}{\longrightarrow} \boldsymbol\vartheta . 
	\end{equation} 
	Because the observed log-likelihoods are continuous functions of $\boldsymbol\vartheta$, the continuous mapping theorem implies
 \begin{equation} \label{eq:map1} 
	\hat{\ell}_{\mathcal X} (\hat{\boldsymbol\vartheta}_N) \overset{P}{\longrightarrow} \ell_{\mathcal X} (\boldsymbol\vartheta),
\end{equation} and, similarly, 
\begin{equation} \label{eq:map2} 
\hat{\ell}_{\mathcal X\setminus \bmX_j} (\hat{\boldsymbol\vartheta}_N) \overset{P}{\longrightarrow} \ell_{\mathcal X\setminus \bmX_j} (\boldsymbol\vartheta). 
\end{equation} Furthermore, by Lemma~\ref{lem:loglikmat}, as the cluster separation tends to infinity, i.e., as $s\to\infty$,
\begin{equation} \label{eq:lem1} 
	\ell_{\mathcal X}(\boldsymbol\vartheta) \longrightarrow l_{\mathcal X}(\boldsymbol\vartheta)
\end{equation} and 
\begin{equation} \label{eq:lem2} 
	\ell_{\mathcal X\setminus \bmX_j}(\boldsymbol\vartheta) \longrightarrow l_{\mathcal X\setminus \bmX_j}(\boldsymbol\vartheta). 
\end{equation} Now, 
\begin{align} 
	& \Big| \hat{\ell}_{\mathcal X\setminus \bmX_j} - \hat{\ell}_{\mathcal X} - \big( l_{\mathcal X\setminus \bmX_j} - l_{\mathcal X} \big) \Big| \nonumber\\ 
	&\qquad\qquad= \Big| \hat{\ell}_{\mathcal X\setminus \bmX_j} - \ell_{\mathcal X\setminus \bmX_j} - \big( \hat{\ell}_{\mathcal X} - \ell_{\mathcal X} \big) + \big( \ell_{\mathcal X\setminus \bmX_j} - l_{\mathcal X\setminus \bmX_j} \big) - \big( \ell_{\mathcal X} - l_{\mathcal X} \big) \Big| \nonumber\\ 
	&\qquad \qquad\le \Big| \hat{\ell}_{\mathcal X\setminus \bmX_j} - \ell_{\mathcal X\setminus \bmX_j} \Big| + \Big| \hat{\ell}_{\mathcal X} - \ell_{\mathcal X} \Big| + \Big| \ell_{\mathcal X\setminus \bmX_j} - l_{\mathcal X\setminus \bmX_j} \Big| + \Big| \ell_{\mathcal X} - l_{\mathcal X} \Big|. 
	\end{align} 
By \eqref{eq:map1} and \eqref{eq:map2}, the first two terms converge to zero in probability as $N\to\infty$. By \eqref{eq:lem1} and \eqref{eq:lem2}, the remaining two terms converge to zero as $s\to\infty$. Therefore, 
\begin{equation} 
	\hat{\ell}_{\mathcal X\setminus \bmX_j} - \hat{\ell}_{\mathcal X} \overset{P}{\longrightarrow} l_{\mathcal X\setminus \bmX_j} - l_{\mathcal X} \quad \text{or, equivalently,} \quad \hat{D}_j \overset{P}{\longrightarrow} D_j
	\end{equation} as $N\to\infty$ and $s\to\infty$. 
	
	Finally, by Theorem~\ref{propmat}, 
	\begin{equation} 
		D_j  \mid (z_{jh}=1) \sim f_{\mathrm{gamma}} \left( d_j-c_h \,\bigg|\, \frac{np}{2},1 \right). 
	\end{equation} 
	Because $c_h$ is  continuous function of $\boldsymbol\vartheta$, consistency of the maximum likelihood estimators and the continuous mapping theorem imply \begin{equation} 
		\hat c_h \overset{P}{\longrightarrow} c_h.
	\end{equation} Hence, 
	\begin{equation} 
		\hat D_j-\hat c_h \overset{P}{\longrightarrow} D_j-c_h
		\end{equation} as $N\to\infty$ and $s\to\infty$,  and the gamma approximation follows: 
		\begin{equation} 
			\hat{\ell}_{\mathcal X\setminus \bmX_j} -\hat{\ell}_{\mathcal X} \mid (z_{jh}=1) \;\dot\sim\; f_{\mathrm{gamma}} \left( \hat d_j-\hat c_h \,\bigg|\, \frac{np}{2},1 \right). 
			\end{equation} 
This completes the proof.

\end{document}